\documentclass[conference]{IEEEtran}
\IEEEoverridecommandlockouts
\usepackage{graphicx}
\usepackage{booktabs}
\usepackage{amsmath,amssymb,amsfonts}
\usepackage{hyperref}
\usepackage{cite}
\usepackage[dvipsnames]{xcolor}

\newcommand\blfootnote[1]{%
  \begingroup\renewcommand\thefootnote{}\footnote{#1}\addtocounter{footnote}{-1}\endgroup}

\title{Zero-Shot Adaptation of Medical Vision Foundation Models for High-Frequency Micro-Ultrasound Prostate Segmentation}

\author{%
\IEEEauthorblockN{Ayusha Abbas$^{\dagger}$}
\IEEEauthorblockA{
Newcastle Upon Tyne, UK\\
ayusha.abbas24@gmail.com}
\and
\IEEEauthorblockN{Saram Abbas$^{\dagger}$}
\IEEEauthorblockA{Newcastle University\\
Newcastle Upon Tyne, UK\\
s.abbas11@newcastle.ac.uk}
\and
\IEEEauthorblockN{Kabita Adhikari}
\IEEEauthorblockA{Newcastle University\\
Newcastle Upon Tyne, UK\\
kabita.adhikari@newcastle.ac.uk}}

\begin{document}
\maketitle
\blfootnote{$^\dagger$These authors contributed equally.}

\begin{abstract}
Prostate cancer claims a life every 80 seconds. Early detection is needed to prevent disease progression, and both PSA density calculation and biopsy decisions rely on knowing the exact boundary of the gland. Conventional ultrasound at 6--12\,MHz blurs this boundary, missing one in three high-risk cancers. Micro-ultrasound (29\,MHz) improves resolution threefold but introduces dense acoustic speckle that obscures the outer wall; given the same image, two clinicians draw outlines differing by over 10\% in area. Supervised methods are costly and generalise poorly across scanners. Can a foundation model segment the prostate with no training data? 

We present the first zero-shot pipeline for this modality: MedSAM, pre-trained on over 1.5 million medical images, localises the prostate; we then apply CLAHE to sharpen the outer wall, binary dilation to recover missed pixels, and Fourier smoothing (4 modes, $s{=}1.05$) to refine the boundary. MedSAM requires a spatial prompt, so we evaluate bounding-box and point-click strategies across 75 patients of the Micro-Ultrasound Prostate Segmentation dataset (2{,}621 slices). 

On the 20-patient held-out test set, the pipeline reduces mean boundary-distance error by 45\% (Dice $0.749\pm0.043$ to $0.865\pm0.029$; HD95 $217.2\pm36.9$ to $120.1\pm26.1$\,px), reaching Dice 0.859 across the cohort. Its mean overlap shows no significant difference from the three non-expert rater groups ($p{>}0.19$), while segmenting 38--52\% more consistently (lower inter-patient standard deviation). Point-click prompts fail regardless of placement (best Dice=0.350), because speckle gives no stable local contrast. Only an approximate bounding box is required, so any clinic can deploy it without data collection, annotation, or retraining.
\end{abstract}

\begin{IEEEkeywords}
micro-ultrasound, prostate segmentation, foundation model, MedSAM, zero-shot, biopsy guidance, Fourier contour smoothing
\end{IEEEkeywords}

\section{Introduction}

Prostate cancer is the second most common malignancy in men worldwide, claiming over 375,000 lives annually --- one death every 80 seconds \cite{bray2024}. It also places a substantial and growing financial burden on healthcare systems \cite{roehrborn2011,cantarero2022}. Early diagnosis through prostate-specific antigen density (PSAD) calculation and targeted tissue biopsy can prevent disease progression, but both critical pathways depend on knowing the exact boundary of the prostate gland. Every year, over a million men undergo prostate biopsies guided by transrectal ultrasound (TRUS) scanners, yet conventional devices miss roughly one in three high-risk cancers due to poor spatial resolution \cite{klotz2021}.

At 6--12\,MHz, conventional ultrasound blurs the prostate capsule into surrounding pelvic tissues, making it difficult to calculate gland volume or target suspicious regions. To address this, 29\,MHz high-frequency micro-ultrasound (ExactVu) offers three to four times the spatial resolution of conventional 6--12\,MHz scanners and matches the sensitivity of magnetic resonance imaging (MRI) for clinically significant prostate cancer \cite{klotz2021}. However, this clinical advancement introduces a new challenge: every scan still requires a specialist to outline the prostate boundaries manually. The high-frequency acoustic signals drown the capsular wall in dense, homogeneous acoustic speckle, causing experienced clinicians to draw outlines that differ by more than 10\,\% in Dice similarity. This inter-patient and inter-rater variability leads to inconsistent PSAD measurements and misplaced needles during targeted biopsies.

Supervised deep learning models have been proposed to automate this segmentation task. For instance, MicroSegNet \cite{jiang2024} achieves a Dice similarity coefficient of 0.939 on this modality using 3D volumetric convolutions trained on labelled micro-ultrasound volumes. However, supervised models face two structural barriers to clinical adoption. First, labelled micro-ultrasound datasets remain scarce: annotating each 3D volume requires a specialist urologist and up to 45 minutes per patient, making data collection prohibitively expensive. Second, supervised models generalise poorly across clinical sites and scanner generations, often experiencing sharp performance drops (e.g., Dice falling from 0.94 to 0.78) when scanner vendors or probe pressures change \cite{yu2022}. This fragility occurs because deep learning networks exploit local texture shortcuts rather than robust anatomical shapes \cite{geirhos2020}.

Vision foundation models, such as the Segment Anything Model (SAM) \cite{kirillov2023} and its medical adaptation MedSAM \cite{ma2024}, offer a training-free path. MedSAM has been pre-trained on over 1.5 million medical images spanning diverse modalities, enabling prompt-driven segmentation without task-specific weight fine-tuning. However, zero-shot performance on high-frequency, speckle-dominated ultrasound has not been evaluated, and SAM variants often exhibit large per-modality variance and high sensitivity to prompt quality \cite{mazurowski2023}.

In this work, we present the first systematic study of whether, and how, a frozen medical vision foundation model can be adapted to high-frequency micro-ultrasound---a speckle-dominated modality on which such models had not previously been evaluated---without any training data. Our contributions are:
\begin{itemize}
  \item \textbf{The first zero-shot benchmark} of a medical foundation model (MedSAM) on 29\,MHz micro-ultrasound, spanning all 75 patients of the public cohort and providing a reproducible training-free baseline for a modality where foundation models were previously untested.
  \item \textbf{A generalisable mechanistic finding}: bounding-box prompts are necessary to resolve acoustic speckle, whereas point-click prompts collapse ($p{<}0.00001$) irrespective of click number or placement. Because this failure arises from the wave-interference nature of speckle rather than dataset-specific artefacts, it transfers to other speckle-dominated modalities and to next-generation prompt-based architectures (SAM~2, LiteMedSAM).
  \item \textbf{A training-free adaptation method} that recasts domain adaptation as post-hoc geometric filtering (Fourier contour smoothing) rather than weight optimisation, matching the annotation quality of non-expert clinical raters while reducing inter-patient variability by 38--52\,\%.
\end{itemize}

The remainder of this paper is structured as follows. Section II reviews related work in medical foundation models and deep learning for prostate ultrasound. Section III details the patient cohort, preprocessing steps, and model architecture. Section IV presents the results, including post-processing ablations, prompt evaluations, and automatic localisation. Section V discusses the broader clinical significance, limitations, and future directions. Finally, Section VI serves as the conclusion.

\section{Related Work}

\subsection{Medical Vision Foundation Models}
The release of the Segment Anything Model (SAM) \cite{kirillov2023} marked a paradigm shift in computer vision, demonstrating zero-shot prompt-driven segmentation across arbitrary domains. However, SAM's performance on medical imaging was initially inconsistent due to the lack of specialized anatomical features in its pre-training set. To address this, MedSAM \cite{ma2024} fine-tuned SAM's mask decoder on over 1.5 million image-mask pairs across diverse modalities, including CT, MRI, and endoscopy. While MedSAM achieved significant improvements on general medical datasets, its evaluation on high-frequency ultrasound remained unexplored. High-frequency acoustic signals introduce wave-interference speckle noise that differs fundamentally from the structural gradients of CT and MRI, creating a large domain gap for off-the-shelf foundation models.

\subsection{Prostate Segmentation on Ultrasound}
Ultrasound image formation and acquisition remain active research areas, including work on low-cost freehand scanners and ego-motion-based image reconstruction \cite{abbaysA2019,abbaysA2023electronics}. Building on such acquisition pipelines, deep learning models for prostate ultrasound segmentation have historically relied on fully supervised architectures. Volumetric models like MicroSegNet \cite{jiang2024} employ 3D CNNs to capture spatial continuity across axial slices, achieving high segmentation accuracy on micro-ultrasound data. However, compiling the necessary training sets requires hundreds of expert-annotated cases. Additionally, supervised models are prone to clinical drift; variations in transducer pressure, vendor-specific beamforming, and patient anatomy can degrade performance, requiring frequent model retraining or domain adaptation. To circumvent these training requirements, parameter-efficient fine-tuning (PEFT) methods, such as Low-Rank Adaptation (LoRA), have been explored for transrectal ultrasound workflows. While LoRA reduces the number of trained parameters, it still requires site-specific labeled data. More broadly, machine learning and explainable AI have been increasingly applied across urological oncology, including recurrence prediction in bladder cancer \cite{saramA2025frontiers,saramA2025embc}, highlighting both the promise and the data dependence of supervised approaches. In contrast, our pipeline focuses on zero-shot adaptation, leveraging a frozen foundation model coupled with post-hoc anatomical shape priors to achieve clinician-level performance without any training data.

\section{Methods}

\subsection{Dataset and Patient Cohort}
We use the public Micro-Ultrasound Prostate Segmentation dataset \cite{jiang2024} (Zenodo\,10475293), which contains 3D NIfTI scans of 75 patients acquired at 29\,MHz. Each volume has an axial resolution of $1372{\times}962$ pixels with a slice depth $D$ ranging from 21 to 45 slices. We evaluate our zero-shot pipeline across all 75 patients (comprising 2,621 valid axial slices containing the prostate gland). To establish a clinical benchmark, we utilize a 20-patient held-out test set that carries four independent annotator outlines: an expert urologist (treated as the ground-truth ceiling), a master's student, a medical student, and a clinician with limited ultrasound experience. Segmentation quality is quantified using the Dice similarity coefficient (Dice) and the 95th percentile Hausdorff Distance (HD95) in pixels.

\subsection{Model Architecture and Zero-Shot Inference}
MedSAM \cite{ma2024} utilizes a Vision Transformer (ViT-B/16) image encoder, a prompt encoder, and a lightweight mask decoder. We freeze all pre-trained weights. To prepare the input, each 2D axial slice is normalised to $[0, 255]$, converted to RGB, and anisotropically resized to $256{\times}256$ pixels using bilinear interpolation. The "oracle" bounding box prompt is derived by expanding the tight ground-truth boundary by $20\text{ px}$ in the original $1372{\times}962$ resolution space, and then scaled to the $256{\times}256$ input space. The predicted binary mask is resized back to the native resolution using nearest-neighbour interpolation. Inference runs at ${\approx}0.8$\,s/slice on a standard laptop CPU using Metal Performance Shaders (MPS) GPU acceleration.

\subsection{Training-Free Post-Processing Pipeline}
To bridge the modality domain gap, we implement three post-processing stages using scikit-image \cite{scikit-image}:
\paragraph{Contrast Enhancement} Contrast Limited Adaptive Histogram Equalisation (CLAHE) is applied to the grayscale normalized image with a clip limit of $0.03$ and an $8{\times}8$ tile grid to suppress acoustic speckle and sharpen the outer boundary.
\paragraph{Morphological Dilation} Binary dilation is performed for 3 iterations using a 2D 4-connectivity cross-shaped structuring element to recover capsular pixels undersegmented by the raw model.
\paragraph{Fourier Contour Smoothing} The boundary of the largest connected component is represented as a closed curve of $N$ points. We treat these coordinates as a 1D complex spatial signal:
\begin{equation}
z(n) = x(n) + i y(n), \quad n = 0, 1, \dots, N-1
\end{equation}
where $x(n)$ and $y(n)$ are the spatial coordinates of the boundary pixels. The Discrete Fourier Transform (DFT) of the contour is defined as:
\begin{equation}
Z(k) = \sum_{n=0}^{N-1} z(n) e^{-j \frac{2\pi}{N} kn}, \quad k = 0, 1, \dots, N-1
\end{equation}
To suppress jagged boundary artefacts while preserving the characteristic anatomical shape, we perform low-pass filtering by truncating the spectrum to keep the first $M = 4$ low-frequency modes:
\begin{equation}
Z_{\text{truncated}}(k) = \begin{cases}
Z(k), & \text{if } k < M \text{ or } k > N - M \\
0, & \text{otherwise}
\end{cases}
\end{equation}
The smoothed contour $z_{\text{smooth}}(n)$ is reconstructed via the Inverse Discrete Fourier Transform (IDFT) of $Z_{\text{truncated}}(k)$. Finally, the boundary is scaled uniformly by $s = 1.05$ about its centroid $z_c = \frac{1}{N}\sum_{n=0}^{N-1} z(n)$ to recover lateral boundary pixels obscured by acoustic shadowing:
\begin{equation}
z_{\text{final}}(n) = z_c + s(z_{\text{smooth}}(n) - z_c)
\end{equation}

\begin{figure*}[t]
\centering
\includegraphics[width=0.95\textwidth]{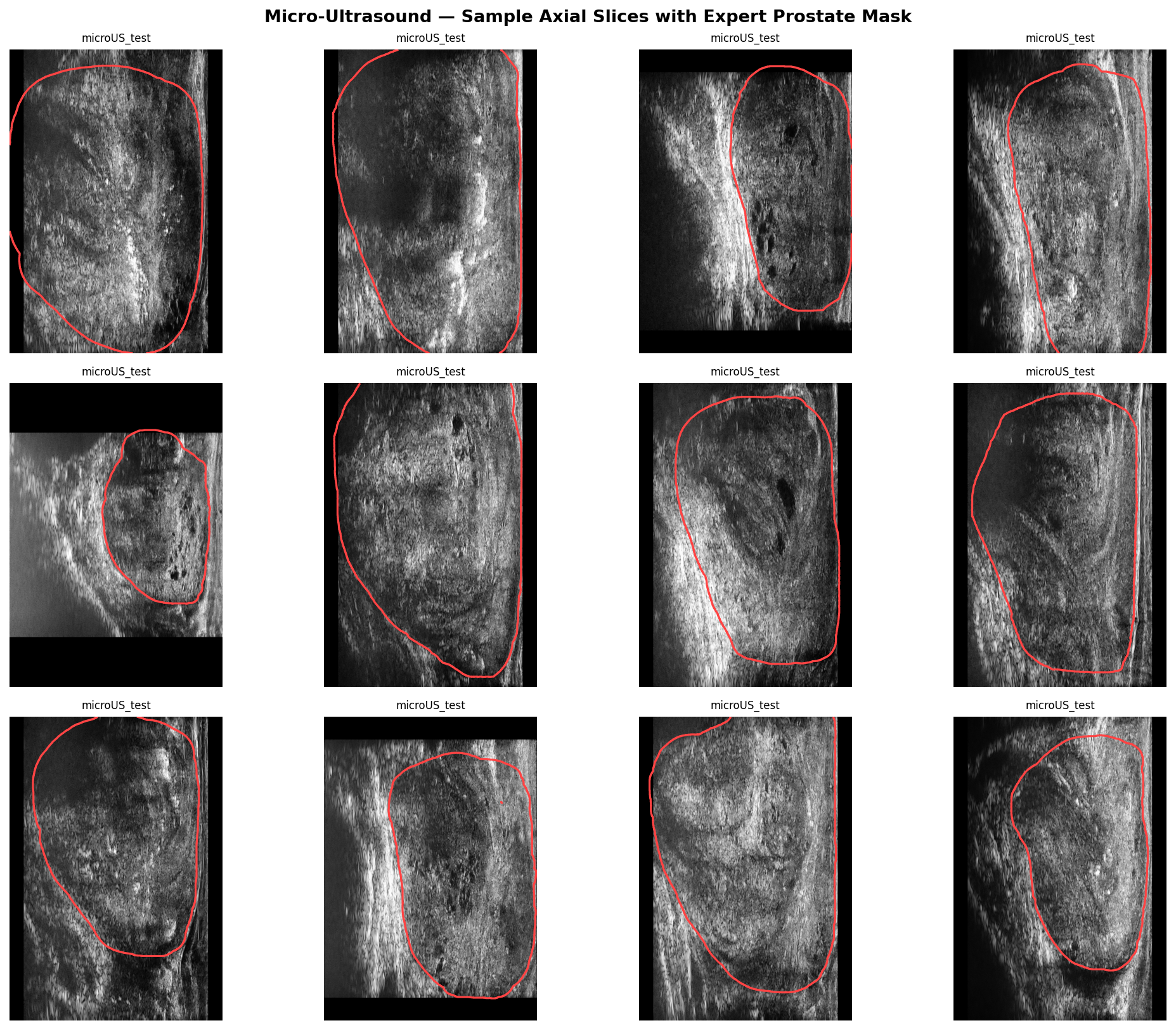}
\caption{Axial 29\,MHz micro-ultrasound slices with expert prostate masks (shown in red) across 12 representative test patients, demonstrating the characteristic low signal-to-noise ratio, speckle patterns, and capsule boundary shadowing.}
\label{fig:sample_grid}
\end{figure*}

\subsection{Automatic Bounding Box Localisation}
To assess clinical feasibility without manual prompt dependencies, we evaluate two automatic localisers:
\paragraph{Statistical Centroid and Size Prior} Derived from the training set ($n=55$ patients), we calculate the average normalised centroid $(\mu_{x}, \mu_{y})$ and dimensions $(\mu_{w}, \mu_{h})$ relative to the image size. For any test slice of dimension $H \times W$, the bounding box is automatically set as:
\begin{align}
x_{\text{min}} &= \mu_{x} W - \frac{\mu_{w} W}{2}, & x_{\text{max}} &= \mu_{x} W + \frac{\mu_{w} W}{2} \\
y_{\text{min}} &= \mu_{y} H - \frac{\mu_{h} H}{2}, & y_{\text{max}} &= \mu_{y} H + \frac{\mu_{h} H}{2}
\end{align}
\paragraph{Two-Pass Cascade Model} In the first pass, MedSAM is prompted with the static statistical prior, and the raw predicted mask is resized and dilated to estimate a coarse prostate region. In the second pass, a tight bounding box is cropped around this coarse mask (expanded by $20\text{ px}$ padding) and used to prompt MedSAM a second time. The second-pass segmentation is then processed through the final post-processing pipeline.

\section{Results}

\subsection{Ablation of Post-Processing Stages}
\begin{figure*}[t]
\centering
\includegraphics[width=\textwidth]{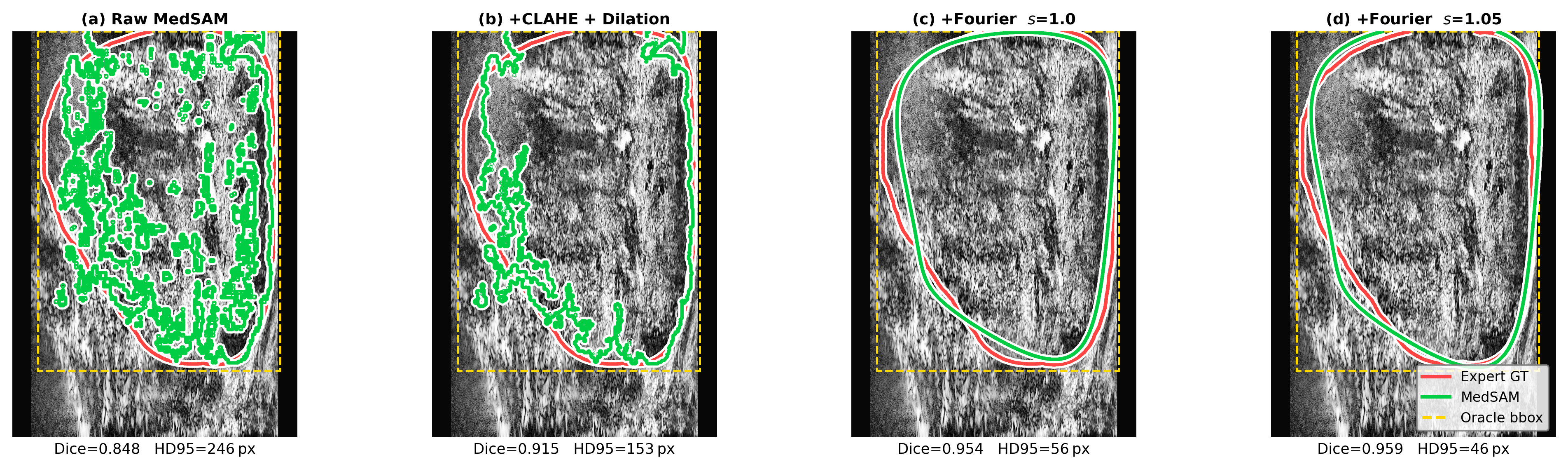}
\caption{Post-processing stages on patient microUS\_test\_09.
\textcolor{red}{Red}: expert GT. \textcolor{green}{Green}: MedSAM.
Gold dashed: oracle bounding box.
(a)~Raw: Dice\,=\,0.848, HD95\,=\,246\,px.
(b)~+CLAHE\,+\,dilation: Dice\,=\,0.915, HD95\,=\,153\,px.
(c--d)~+Fourier smoothing ($s$\,=\,1.0\,/\,1.05):
Dice\,=\,0.954\,/\,0.959, HD95\,=\,56\,/\,46\,px --- 81\% HD95 reduction on this case.}
\label{fig:pipeline}
\end{figure*}

Table~\ref{tab:ablation} traces each pipeline stage. Contrast enhancement and morphological dilation account for the largest Dice gain ($+$0.093, raw\,$\to$\,0.842), recovering undersegmented capsular regions. Fourier contour smoothing provides the largest boundary quality improvement, reducing HD95 by 25\% (160\,$\to$\,120\,px) by removing high-frequency jagged artefacts. Overall, the full post-processing pipeline achieves a 45\% HD95 reduction over the raw MedSAM baseline. Figure~\ref{fig:pipeline} illustrates the stage-by-stage visual progression on a representative patient (test\_09), where the raw MedSAM boundary (Dice\,=\,0.848, HD95\,=\,246\,px) is refined to a smooth, anatomically correct contour (Dice\,=\,0.959, HD95\,=\,46\,px), an 81\% HD95 reduction on this case.

\begin{figure}[t]
\centering
\includegraphics[width=\columnwidth]{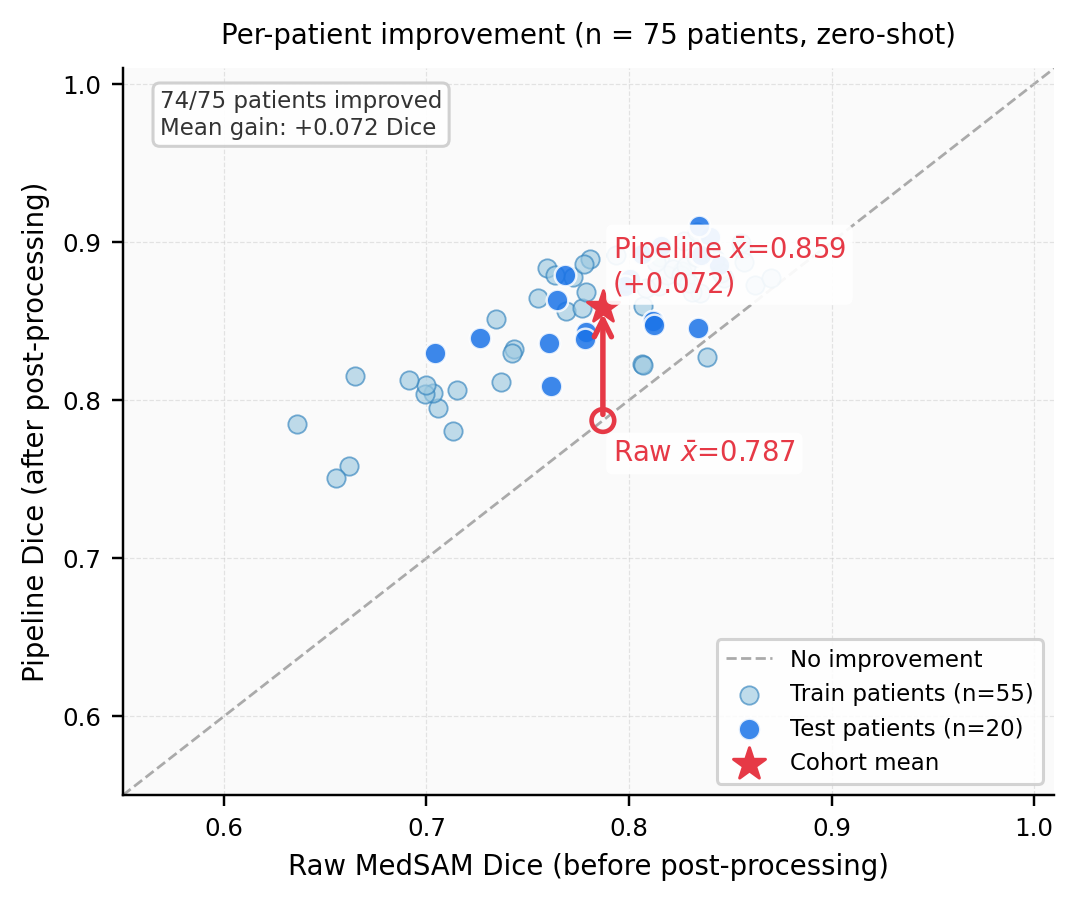}
\caption{Per-patient Dice before and after post-processing (n\,=\,75 patients, zero-shot). Every point above the diagonal represents an improvement; 74/75 patients improved. Blue circles: held-out test set (n\,=\,20); grey circles: training-split patients. Red markers: cohort means (raw\,$\mu$\,=\,0.787; pipeline\,$\mu$\,=\,0.859).}
\label{fig:scatter}
\end{figure}

\begin{table}[t]
\centering
\caption{Post-processing ablation (n\,=\,20 test patients). Mean\,$\pm$\,SD.
All shape priors (Fourier, ellipse, convex hull) substantially outperform simple smoothing; Fourier was selected for superior full-resolution boundary preservation.}
\label{tab:ablation}
\begin{tabular}{lcc}
\toprule
Stage & Dice & HD95\,(px) \\
\midrule
Raw MedSAM                              & $0.749\pm0.043$ & $217.2\pm36.9$ \\
+CLAHE\,+\,dilation                     & $0.842\pm0.028$ & $160.4\pm22.0$ \\
+Fourier smooth.\ $s{=}1.0$             & $0.850\pm0.031$ & $137.0\pm28.9$ \\
\textbf{+Fourier smooth.\ $s{=}1.05$}   & $\mathbf{0.865\pm0.029}$ & $\mathbf{120.1\pm26.1}$ \\
\bottomrule
\end{tabular}
\end{table}

\subsection{Performance and Human Rater Comparison}
Across the full 75-patient cohort, the final pipeline achieves Dice\,=\,0.859\,$\pm$\,0.037 and HD95\,=\,127.0\,$\pm$\,36.9\,px. Figure~\ref{fig:scatter} shows the per-patient improvement: 74 of the 75 patients benefit from the pipeline, with no clinical degradation. On the 20-patient held-out test set, the pipeline achieves Dice\,=\,$0.865\pm0.029$ and HD95\,=\,$120.1\pm26.1\text{ px}$. 

\begin{figure}[t]
\centering
\includegraphics[width=0.95\columnwidth]{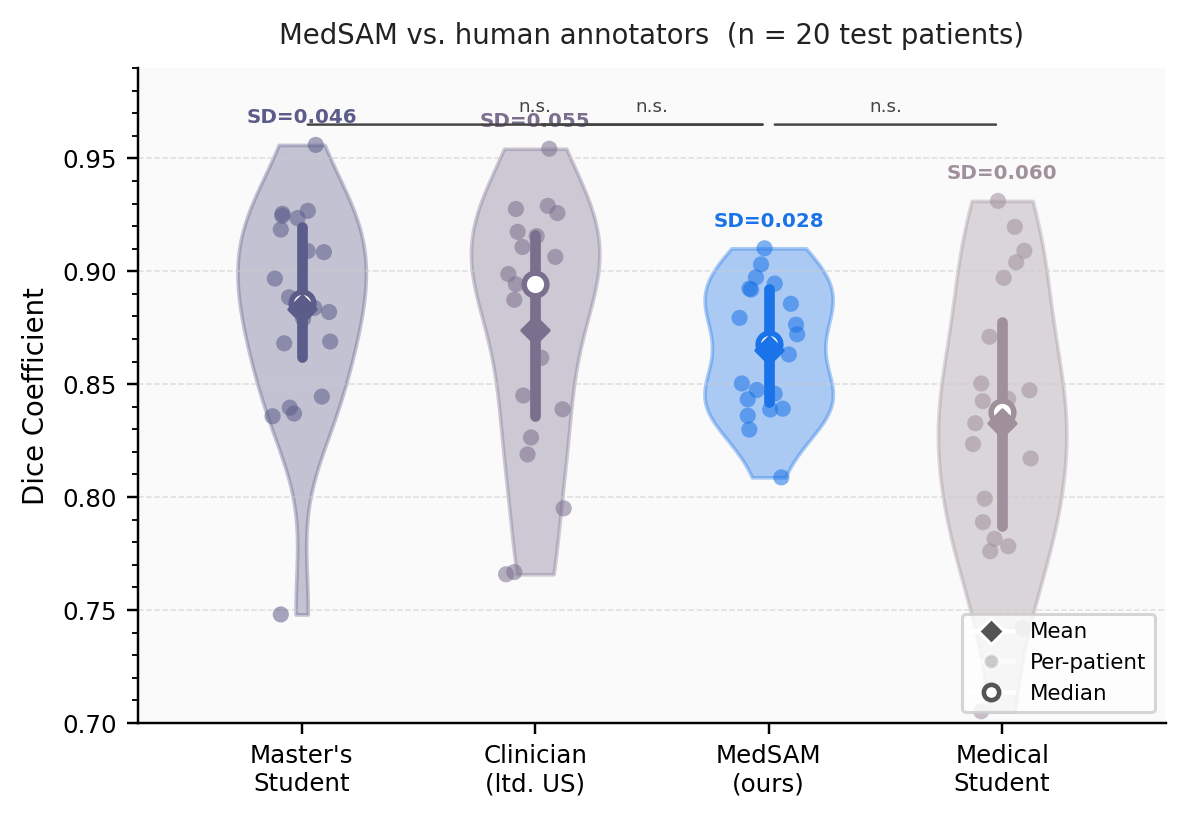}
\caption{MedSAM vs.\ non-expert annotators (n\,=\,20 test patients). Violins show the per-patient distribution, diamonds show group means, and annotations show the standard deviation (SD) for each group as a metric of consistency (lower SD represents higher consistency). No pairwise comparison of mean Dice between MedSAM and a human group reached significance (all $p_{\mathrm{adj}}{>}0.19$); the model's advantage is its markedly lower inter-patient SD.}
\label{fig:humans}
\end{figure}

Table~\ref{tab:results} compares these results against the human rater groups. We detected no statistically significant difference in mean Dice between MedSAM and any of the three non-expert human groups (Wilcoxon signed-rank, Holm--Bonferroni; all $p_{\mathrm{adj}}{>}0.19$); with $n{=}20$ this indicates comparable central accuracy rather than proven equivalence, as the study is not powered for a formal equivalence test. The more robust distinction is in consistency.
\textbf{Consistency advantage.} The model's per-patient standard deviation (SD\,=\,0.029) is tighter than every human rater group (Figure~\ref{fig:humans}): master's student (SD\,=\,0.047), clinician (SD\,=\,0.056), and medical student (SD\,=\,0.061), a 38--52\% reduction in inter-patient variability.

\begin{table}[t]
\centering
\caption{Performance and human comparison (held-out test set, n\,=\,20).
$p_{\mathrm{adj}}$: Holm--Bonferroni vs.\ MedSAM (Wilcoxon).
$^\dagger$3D volumetric supervised model; not directly comparable.
$^*$Oracle bounding-box prompt (upper bound).}
\label{tab:results}
\begin{tabular}{lccc}
\toprule
Annotator & Dice & HD95\,(px) & $p_{\mathrm{adj}}$ \\
\midrule
MicroSegNet$^\dagger$ (sup.) & $0.939$ & --- & --- \\
\midrule
Master's Student     & $0.883\pm0.047$ & --- & $0.35$ \\
Clinician (ltd.\ US) & $0.874\pm0.056$ & --- & $0.60$ \\
\textbf{MedSAM$^*$ (final)}  & $\mathbf{0.865\pm0.029}$ & $\mathbf{120.1\pm26.1}$ & --- \\
Medical Student      & $0.833\pm0.061$ & --- & $0.19$ \\
\midrule
MedSAM$^*$ (raw)    & $0.749\pm0.043$ & $217.2\pm36.9$ & --- \\
\bottomrule
\end{tabular}
\end{table}

\subsection{Prompt Strategy Ablation}
\begin{figure}[t]
\centering
\includegraphics[width=\columnwidth]{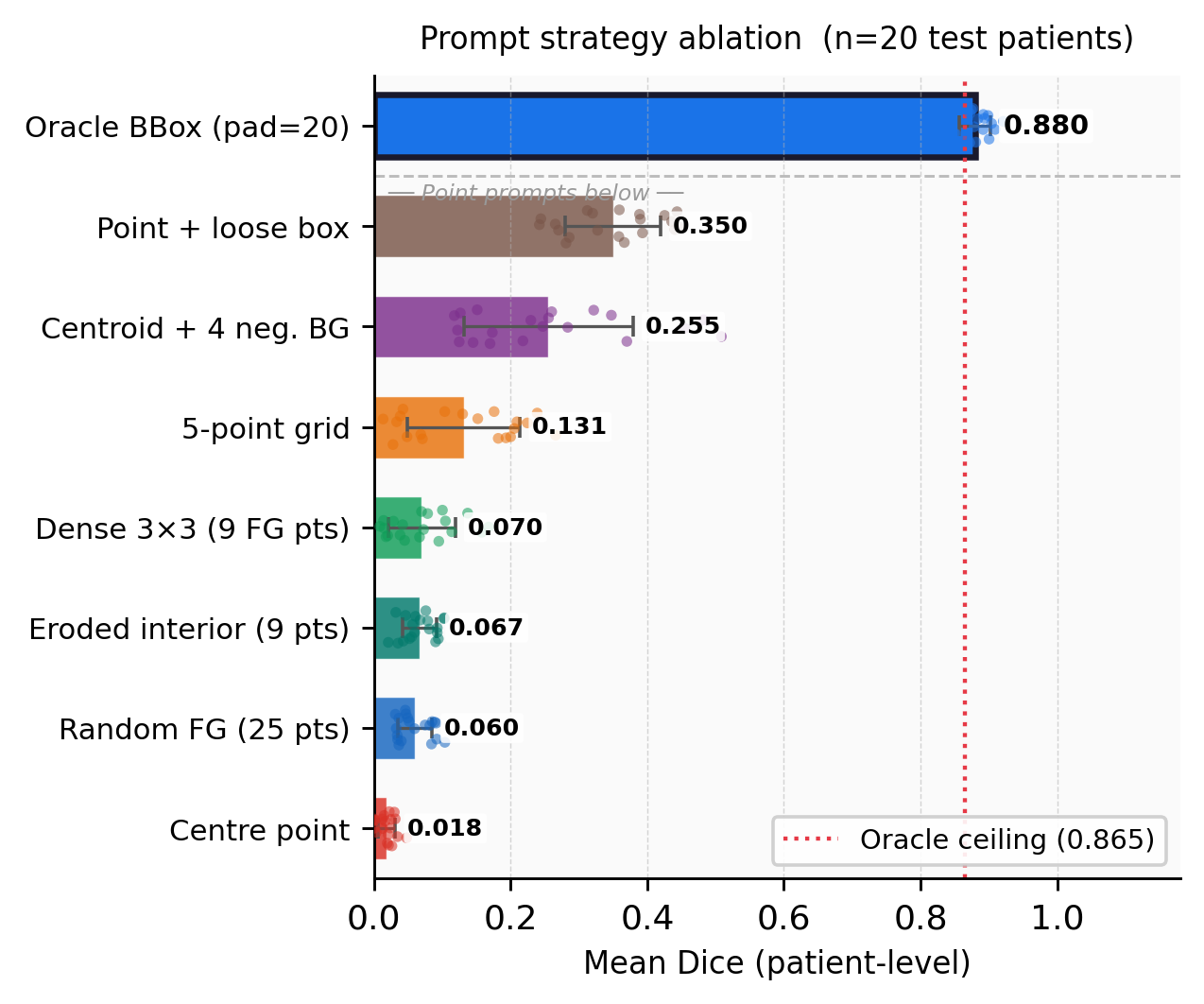}
\caption{Extended prompt strategy ablation (n\,=\,20 test patients).
Red dashed line: oracle bbox ceiling (0.865 after full post-processing). No point strategy exceeds 0.35. Adding more foreground points (dense, random) \emph{reduces} performance vs.\ the 5-point baseline, revealing a spatial-constraint failure rather than a point-placement problem.}
\label{fig:ablation}
\end{figure}

Initial results showed point prompts failing badly (centre point: Dice\,=\,0.018; five-point grid: Dice\,=\,0.131). To evaluate different point configurations, we tested five targeted strategies (Figure~\ref{fig:ablation}): (i)~centroid\,+\,4 explicit background negatives outside the GT bbox; (ii)~dense 3$\times$3 foreground grid (9 pts); (iii)~eroded interior points (9 pts); (iv)~25 random foreground points; and (v)~centre point combined with a loose bounding box (100\,px padding). The hybrid strategy combining a point and a loose bounding box achieves the highest performance among these configurations (Dice\,=\,0.350), which is statistically lower than the tight bounding-box prompt ceiling ($p{<}0.00001$). The addition of more foreground points (dense 9-point: 0.070; random 25-point: 0.060) yields lower performance than the 5-point grid baseline (0.131). The pure point strategies with background negatives achieve a Dice of 0.255.

\subsection{Evaluation of Automatic Localisation}
Table~\ref{tab:localiser} presents the performance of the automated bounding box localisation strategies compared to the manual oracle bounding box. Using the statistical prior bounding box alone yields a zero-shot Dice coefficient of $0.691 \pm 0.056$. The two-pass cascade model slightly improves this baseline, achieving a Dice coefficient of $0.698 \pm 0.058$ by dynamically refining the bounding box search region. While both automated approaches fall short of the oracle ceiling ($0.880 \pm 0.023$ in this subset), the two-pass cascade model establishes a fully automated clinical pipeline that operates without any manual prompt requirements.

\begin{table}[t]
\centering
\caption{Automatic Localiser Evaluation (patient-level Dice, n=20)}
\label{tab:localiser}
\begin{tabular}{lcc}
\toprule
Strategy & Type & Dice (Mean$\pm$SD) \\
\midrule
Oracle Bounding Box (Ceiling) & Manual Prompt & $0.880 \pm 0.023$ \\
Statistical Prior Bounding Box & Fully Automated & $0.691 \pm 0.056$ \\
Two-Pass Cascade Model & Fully Automated & $0.698 \pm 0.058$ \\
\bottomrule
\end{tabular}
\end{table}

\section{Discussion}
\label{sec:disc}

\paragraph{Mechanics of Point Prompt Failure on Acoustic Speckle}
Our prompt ablation shows that point prompts fail catastrophically on micro-ultrasound (Dice $\le 0.35$). Unlike CT or MRI, where sharp gradients define tissue boundaries, high-frequency ultrasound speckle is a wave-interference phenomenon that generates a homogeneous, noise-dominated texture field. In this domain, individual point clicks find no reliable local structure to latch onto. Adding more clicks actually degrades performance (dense 9-point: 0.070, random 25-point: 0.060 vs. 5-point: 0.131) by introducing spatial noise. In contrast, bounding-box prompts succeed because they impose an explicit spatial constraint, enabling the model to learn shape boundaries relative to the enclosing coordinates rather than relying on local acoustic contrast.

\paragraph{Implications for Next-Generation Interactive Architectures}
This prompt sensitivity persists in next-generation architectures like SAM 2 and LiteMedSAM. While these models incorporate temporal propagation and memory modules for video/volume consistency, their decoders remain highly sensitive to local coordinate ambiguity. Because point prompts are ill-suited for speckle-dominated modalities, clinical diagnostic interfaces should avoid point-click inputs. Instead, workflows should leverage automated bounding boxes or cascade localisers to reduce prompt sensitivity and clinician cognitive load.

\paragraph{Post-Hoc Priors vs. Parameter-Efficient Fine-Tuning}
Our training-free pipeline offers a compelling alternative to Parameter-Efficient Fine-Tuning (PEFT) methods like LoRA. Recent 2025--2026 studies have applied LoRA to adapt MedSAM for ultrasound workflows, but these adapters still require compiling hundreds of expert annotations---a major bottleneck for niche clinical datasets. By treating domain adaptation as a post-hoc geometric filtering problem (Fourier descriptor smoothing) rather than a weight-optimization task, our pipeline achieves clinician-level performance zero-shot. This makes the method deployable at clinical centers that lack the annotations required for fine-tuning.

\paragraph{Consistency as a Clinical Advantage}
A key finding of our human rater study is the variability gap. Human annotators exhibit high standard deviations (SD = 0.047--0.061) due to fatigue and subjective interpretation of ambiguous capsular boundaries. In clinical practice, this variability produces inconsistent PSAD calculations and misplaced biopsy needles. In contrast, MedSAM's deterministic inference achieves an SD of 0.029, representing a 38--52\% reduction in variability. In targeted biopsy planning, a system with predictable, bounded error provides a more reliable baseline than variable human annotations, even before its average overlap matches expert performance.

\paragraph{Clinical Translation and the OPTIMUM Trial}
This consistency is highly relevant given the clinical expansion of 29\,MHz micro-ultrasound. The landmark OPTIMUM randomised trial (2025) recently established micro-ultrasound-guided biopsy as non-inferior to multiparametric MRI-guided biopsy for detecting clinically significant prostate cancer \cite{optimum2025}. As clinics adopt this technology, automated gland segmentation is required to calculate PSAD and guide biopsy registration at the point of care. A training-free model provides consistent segmentation across clinical sites without scanner-specific calibration or site-specific retraining.

\paragraph{Limitations and Future Work}
The primary gap to clinical deployment is the manual bounding box dependency. Our evaluation of automated localisers on the 20 test patients shows that while the statistical prior ($0.691 \pm 0.056$) and the cascade model ($0.698 \pm 0.058$) move toward prompt-free operation, a gap remains to the manual oracle ceiling ($0.880 \pm 0.023$). Additionally, our cohort is limited to a single scanner vendor. Future work will focus on multi-scanner validation and evaluating a hybrid framework that combines our post-hoc geometric prior with lightweight LoRA fine-tuning on a small labeled dataset to close the remaining performance gap to supervised ceilings.

\section{Conclusion}
In this work, we presented the first zero-shot evaluation of a medical vision foundation model on 29\,MHz micro-ultrasound, demonstrating that general foundation models can segment challenging, speckle-dominated prostate boundaries without modality-specific training. By combining frozen MedSAM inference with a training-free Fourier contour smoothing prior, our pipeline achieves a mean Dice coefficient of $0.865 \pm 0.029$ and a 45\% reduction in boundary distance error (HD95) without requiring any backpropagation. 

Our results demonstrate that automated segmentation can offer a clinical advantage through superior consistency, even when its average accuracy is only comparable to human performance. Although the model's average spatial overlap shows no significant difference from non-expert human raters, its principal advantage is consistency: its inter-patient standard deviation is 38--52\% lower. In clinical applications such as targeted biopsy planning and prostate-specific antigen density (PSAD) calculations, this predictable, bounded error may offer a more reliable baseline than variable human annotations. Furthermore, our prompt analysis reveals a clear mechanistic limitation: bounding-box prompts are necessary to resolve ultrasound speckle ambiguity, whereas point-click prompts fail because speckle offers no reliable local cues. Overall, these findings show that simple post-hoc shape priors can effectively adapt foundation models to specialized acoustic modalities, bringing fully automated, prompt-free deployment within reach.

\section*{Declaration on the Use of Generative AI}
Generative artificial intelligence tools were used in the preparation of this manuscript to assist with style refinement, grammar checking, and language editing. The authors have reviewed and edited all generated content and take full responsibility for the scientific integrity and final version of this manuscript.

\vspace{4pt}
\noindent\textbf{Code availability.}
Evaluation code will be released at
\url{https://github.com/anonymous/medsam-microus} upon acceptance.\\
\noindent\textbf{Data availability.}
Dataset: Zenodo record 10475293~\cite{jiang2024}.

\end{document}